\documentclass[journal]{IEEEtran}
\usepackage[english]{babel}
\usepackage[utf8]{inputenc}

\usepackage[capitalise]{cleveref}
\usepackage{blindtext}
\usepackage{graphicx}
\usepackage{listings}
\usepackage{caption}
\usepackage{xcolor}
\usepackage{geometry}
\usepackage{DejaVuSans}
\usepackage{DejaVuSansMono}
\usepackage{amssymb}
\usepackage{cleveref}
\usepackage{amsmath}
\usepackage{multirow}
\usepackage{amsmath}
\usepackage[normalem]{ulem}

\usepackage{hyperref}
\hypersetup{
    colorlinks=true,
    linkcolor=blue,
    filecolor=magenta,      
    urlcolor=cyan,
    citecolor=blue,
}

\usepackage{algorithm}
\usepackage{algorithmic}

\DeclareCaptionFormat{listing}{\rule{\dimexpr0.9\columnwidth+17pt\relax}{0.4pt}\par\vskip1pt#1#2#3}
\DeclareMathVersion{sans}
\SetSymbolFont{operators}{sans}{OT1}{cmbr}{m}{n}
\SetSymbolFont{letters}  {sans}{OML}{cmbrm}{m}{it}
\SetSymbolFont{symbols}  {sans}{OMS}{cmbrs}{m}{n}

\lstnewenvironment{sflisting}[1][]
  {\lstset{#1}\mathversion{sans}}{}

\begin{document}
\title{ROS2Learn: a reinforcement learning framework for ROS 2}


%
%
%


\author{ Yue Leire Erro Nuin$^{*}$, Nestor Gonzalez Lopez$^{*}$, Elias Barba Moral$^{*}$, Lander Usategui San Juan, Alejandro Solano Rueda, 
Víctor Mayoral Vilches and Risto Kojcev\\
\emph{Acutronic Robotics, March 2019}
\thanks{$^{*}$ authors contributed equally}%
}

\maketitle

\begin{abstract}
We propose a novel framework for Deep Reinforcement Learning (DRL) in modular robotics to train a robot directly from joint states, using traditional robotic tools. We use an state-of-the-art implementation of the Proximal Policy Optimization, Trust Region Policy Optimization and Actor-Critic Kronecker-Factored Trust Region algorithms to learn policies in four different Modular Articulated Robotic Arm (MARA) environments. We support this process using a framework that communicates with typical tools used in robotics, such as Gazebo and Robot Operating System 2 (ROS 2). We evaluate several algorithms in modular robots with an empirical study in simulation.
\end{abstract}

\IEEEpeerreviewmaketitle

\section{Introduction}

Current robot systems are designed, built and programmed by teams with multidisciplinary skills. The traditional approach to program such systems is typically referred to as the \emph{robotics control pipeline} and requires going from observations to final low-level control commands through: a) state estimation, b) modeling and prediction, c) planning, and d) low level control translation \cite{mayoral2018modular}. As introduced by Zamalloa et al. \cite{zamalloa2017dissecting}, the entire process requires the fine tuning of every step in the pipeline, incurring in a significant complexity, where optimization at every step is critical and has a direct impact in the final result.

Artificial Intelligence methods and, particularly, neuromorphic techniques such as artificial neural networks (ANNs) are becoming more and more relevant in robotics. Starting from 2016, promising results, such as the work of Levine et al. \cite{DBLP:journals/corr/LevinePKQ16}, showed a path towards simplifying the construction of robot behaviours through the use of deep neural networks as a replacement of the traditional approach outlined above. The described end-to-end approach for programming robots scales nicely when compared to traditional methods.

Reinforcement Learning (RL) is a field of machine learning that is concerned with making sequences of decisions. It considers an agent situated in an environment where for each timestep the agent takes an action and receives an observation and a reward. A RL algorithm seeks to maximize the agent's total reward trough a trial and error learning process. Deep Reinforcement Learning (DRL) is the study of RL by using neural networks as function approximates. In recent years, several techniques for DRL have shown good success in learning complex behaviour skills and solving challenging control tasks in high-dimensional state-space \cite{levine2013guided, peters2008reinforcement, schulman2015trust, schulman2017proximal, wu2017scalable}. However, many of the benchmarked environments such as Atari \cite{mnih2013playing} and Mujoco \cite{todorov2012mujoco} rarely deal with realistic or complex environments (frequent in robotics) \cite{7943603, zamora2016extending}, or use the tools commonly used in the field such as the Robot Operating System (ROS)\cite{quigley2009ros}. The research conducted in the previous work can only be translated into real world robots with a considerable amount of effort for each particular robot. Hence, the scalability of previous methods for modular robots is questionable. 

Modular robots can extend their components seamlessly by just adding modules to the robotic system. This brings clear advantages for the construction of robots, however training them with current DRL methods becomes cumbersome due to the following reasons: every small change in the physical structure of the robot will require a new training; building the tools to train modular robots (such as the simulation model, virtual drivers) is a time consuming process; transferring the results to the real robot is complex given the flexibility of these systems. In this work we present a framework that utilizes the traditional tools in the robotics environment, such as Gazebo\cite{koenig2004design} and ROS 2, which simplifies the process of building modular robots and their corresponding tools. Our framework includes baseline implementations\cite{baselines} for the most common DRL techniques for policy iteration methods. Using this framework we present the results obtained benchmarking DRL methods in a modular robot with 6 degrees-of-freedom (DoF).

\section{Previous Work}

Recent advances in the field of RL have led to the development of different approaches with neural network function approximators. Among the available techniques, the focus of this work is on model-free RL methods: Proximal Policy Optimization (PPO) \cite{schulman2017proximal} and natural gradient policy based methods, such as Trust Region Policy Optimization (TRPO) \cite{schulman2015trust} and Actor Critic using Kronecker-Factored Trust Region (ACKTR) \cite{wu2017scalable}. All of these are known as policy gradient methods, which perform updates at each episode to the policy parameters (on-policy).

\textbf{TRPO} \cite{schulman2015trust} is a policy gradient method meant to solve RL problems more efficiently than "vanilla" policy gradient (VPG) \cite{Sutton:1999}. The idea used is to update the weights as fast as possible without diverging. For achieving this, TRPO uses a constrain linked to the KL-Divergence \cite{kullback1951information}, that gives a measure of distance between two probability distributions. TRPO can be applied both for learning non-trivial tasks in continuous control as well as for discrete control policies directly from raw pixel inputs. Thanks to the use of the natural policy gradient method, TRPO overcomes some limitations of VPG such as choosing the step-size and the low sample efficiency. Compared to other algorithms \cite{duan2016benchmark} \cite{ghadir2017trpo}, TRPO has proven to be a good approach for continuous control tasks. Details of the theoretical aspects of the TRPO method are given in Section \ref{methods:trpo}.

\textbf{ACKTR} is an actor-critic RL method that applies trust region policy optimization using Kronecker-factored approximation (K-FAC) to the curvature. This method uses the natural policy gradient and optimizes both the actor and the critic \cite{wu2017scalable}. Similarly to TRPO, ACTKR also uses the benefits from natural policy gradient and can be applied both in continuous and discrete environments. In the evaluation of Wu et. al \cite{wu2017scalable}, ACKTR sample and computational efficiency was evaluated in Atari and several Mujoco environments, and it was compared with the performance of Advantage Actor Critic (A2C) and TRPO. Wu et. al \cite{wu2017scalable} results indicate that the performance of ACKTR surpassed the performance of A2C and TRPO. In this work, we extend the evaluation of ACKTR to a set of environments which are particular for representing different robot configurations and scenarios. Details of the theoretical aspects of the ACKTR method are given in Section \ref{methods:acktr}. 

\textbf{PPO} is a policy gradient method for RL which alternates between sampling data trough interaction with the environment and optimizing the "surrogate" objective using Stochastic Gradient Descent \cite{schulman2017proximal}. PPO differs from standard policy gradient methods by enabling multiple epochs of mini batch updates. Compared to its predecessor (TRPO), PPO uses the "surrogate" objective by clipping the policy probability ratio. In the original work, PPO was evaluated in the Atari, Mujoco and Roboschool \cite{roboschool} environments, where it had better performance compared to A2C, A2C + Trust region, VPG and TRPO. Details of the theoretical aspects of PPO are given in Section \ref{methods:ppo}.

Previous works, \cite{barrett2010transfer, DBLP:journals/corr/RusuVRHPH16, DBLP:journals/corr/JamesJ16}, present partial success of transferring learned behaviour in simulation to a real robot. These works explain the importance of having scenes in simulation as similar as possible to the reality in order to simplify the process of transferring the learned behaviour to real scenarios. Yuke Zhu et. al \cite{zhu2017target} describe high-quality and realistic 3D scenes. The approach of Tobin et. al \cite{tobin2017domain}, randomizes the rendering in simulation, reaching enough variability. This allows for the images in the real world to be considered as just another variation in the simulator. To the best of our knowledge, the work conducted in previous approaches focuses on restricted scenarios in a controlled environment where specific algorithms for solving particular task were used. This is not the case when a robotic system needs to be deployed in realistic scenarios, specially if the robot is modular and can present a number of different configurations.

The methods presented above have all different theoretical approaches for solving RL tasks with their strengths and drawbacks, which makes it hard to determine which one is the most appropriate choice for a particular application and environment. The aim of this work is to evaluate the above mentioned RL algorithms with the focus of determining which one of them is best suited for modular robotic applications. Section \ref{methods} describes the theoretical aspects of the evaluated RL methods and their adaptation to applications for modular robots. Section \ref{experiments} presents the experimental evaluation conducted in 6DoF modular Modular Articulated Robotic Arm (MARA) robot\footnote{\url{https://acutronicrobotics.com/products/mara/}}. Section \ref{conclusion} summarizes results and presents future perspective and work.

\section{Methods}
\label{methods}

\subsection{Nomenclature}
\label{methods:nomenclature}

The methods presented bellow will be consistent with the following nomenclature that is partially inspired on the work by Peters et al. \cite{Peters:2010}:

The three main components of a RL system for robotics include the state $s$ (also found in literature as $x$), the action $a$ (also found as $u$) and the reward denoted by $r$. We will denote the current time step by $k$. The stochasticity of the environment gets represented by using a probability distribution $\mathbf{s}_{k+1}\sim p\left(  \mathbf{s}_{k+1}\left\vert \mathbf{s}_{k},\mathbf{a}_{k}\right.  \right)$ as model where $\mathbf{a}_{k}\in\mathbb{R}^{M}$ denotes the current action and $\mathbf{s_k}$, $\mathbf{s}_{k+1}\in\mathbb{R}^{N}$ denote the current and next state, respectively. Further, we assume that most policy gradient methods have actions that are generated by a policy $\mathbf{a}_{k} \sim \mathbf{\pi}_{\mathbf{\theta}}\left( \mathbf{a}_{k}\left\vert \mathbf{s}_{k}\right.  \right)$ which is modeled as a probability distribution in order to incorporate exploratory actions.

The policy is assumed to be parametrized by $K$ policy parameters $\mathbf{\theta} \in\mathbb{R}^{K} $. The sequence of states and actions forms a trajectory denoted by $\mathbf{\tau}=[\mathbf{s}_{0:H},\mathbf{a}_{0:H}]$ where $H$ denotes the horizon which can be infinite. Often, \emph{trajectory}, \emph{history}, \emph{trial} or \emph{roll-out} are used interchangeably. At each instant of time, the learning system receives a reward $r_{k} = r\left(  \mathbf{s} _{k},\mathbf{a}_{k}\right)
\in\mathbb{R} $.

The general goal of policy optimization is to optimize the policy parameters $\mathbf{\theta}\in\mathbb{R}^{K}$ so that the expected return is optimized:
\begin{equation}
J\left(  \mathbf{\theta}\right)  =E\left\{  \sum\nolimits_{k=0}^{H} \mathbf{\gamma} \cdot r_{k}\right\}
\end{equation}
where $ \mathbf{\gamma} \in [0,1]$ is the discount factor.

For real-world applications, we require that any change to the policy parameterization has to be smooth. Otherwise, drastic changes can be hazardous for the actor, and useful initializations of the policy based on domain knowledge would vanish after a single update step. For these reasons, policy gradient methods which follow the steepest descent on the expected return are the method of choice. These methods update the policy parameters according to the gradient update rule
\begin{equation}
 \mathbf{\theta}_{h+1}=\mathbf{\theta}_{h}+ \mathbf{\alpha}_{h}\left.  \mathbf{\nabla}_{\mathbf{\theta}}J\right\vert _{\mathbf{\theta}=\mathbf{\theta}_{h}},
\end{equation}
where $ \mathbf{\alpha}_{h}\in\mathbb{R}^{+}$ denotes the learning rate and $h\in\{0,1,2,\ldots\}$ the current update number.\\

\begin{table}[h!]
    \small
    \centering
    \caption{Summary of the terms used within the article.}
    \begin{tabular}{ c  p{3cm} }
        $s$ & the \emph{state} (also found in literature as $x$) \\ \hline
        $a$ & the \emph{action} (also found as $u$) \\ \hline
        $r$ & the \emph{reward} \\ \hline 
        $k$ & time step \\ \hline
        $\mathbf{s}_{k+1}\sim p\left(  \mathbf{s}_{k+1}\left\vert \mathbf{s}_{k},\mathbf{a}_{k}\right.  \right)$ &  probability distribution representing the stochasticity of the environment \\ \hline
        
        $ \mathbf{\gamma}$ & discount factor \\ \hline
        $\tau$ & trajectories \\ \hline
        $\mathbf{\tau}\sim p_{\mathbf{\theta}}\left(  \mathbf{\tau}\right) =p\left(  \left. \mathbf{\tau}\right\vert \mathbf{\theta}\right)$ & roll-outs \\ \hline
        $r(\mathbf{\tau})=\sum\textstyle_{k=0}^{H} \mathbf{\gamma} r_{k}$ & return in the roll-outs \\
    \end{tabular}
    \label{table:terms}
\end{table}

\subsection{Benchmarked algorithms}
One of the main distinctions between algorithms in RL is based on if they are value-based or policy-based. The first class, value-based, attempts to learn to assess correctly what is the reward obtained in a certain state and thus, maximize the final expected reward. The second class, policy-based, attempts to learn what action to do at each state in order to maximize the final reward. Robotics is dominated by scenarios with continuous changes in states and actions spaces, which implies that most traditional value-based off-the-shelf RL approaches are not valid for treating such situations. As pointed out by Peters et al. \cite{Peters:2010}, Policy Gradient (PG) methods differ significantly from others as they do not suffer from these problems in the same way other techniques do.\\
One of the typical problems experienced when doing RL in robotics\footnote{provided that there is no additional state estimator (actor-critic methods)} is that uncertainty in the state might degrade the performance of the policy. PG methods suffer from this as well. However, they rely on optimization techniques for the policy that do not need to be changed when dealing with this uncertainty.\\
The nature of PG methods allows them to deal with continuous states and actions in exactly the same way as discrete ones. PG techniques can be used either on model-free or model-based approaches. The policy representation can be chosen in order to be meaningful for the task, and can incorporate domain knowledge. This often leads to the use of fewer parameters in the learning process. Additionally, its generic formulation shows that PG methods are valid even when the reward function is discontinuous or even unknown.

While PG techniques might seem interesting for a roboticist on a first look, they are by definition on-policy and need to forget data reasonably fast in order to avoid the introduction of a bias to the gradient estimator. In other words, they are not as good as other techniques at using the data available (their sample efficiency is low). Other typical problem with PG methods is that convergence is only guaranteed to a local maximum while in tabular representations, value function methods are guaranteed to converge to a global maximum.\\

\subsubsection{Trust Region Policy Optimization (TRPO)}
\label{methods:trpo}
Trust Region Policy optimization is an attempt of improving VPG, by choosing appropriately the magnitude of update at each iteration \cite{schulman2015trust}. In order to know how much to update, TRPO uses the KL-divergence, which returns a measure of how different two probability distributions are. The formal definition of the problem is:
\begin{equation}
\begin{aligned}
\mathbf{\theta}_{k+1} = & arg\,\max\limits_{\mathbf{\theta}}L_{\mathbf{\theta}}(\mathbf{\theta}_{k})\\
&s.t. D_{KL}(\mathbf{\theta}\|\mathbf{\theta}_{k})\leq \mathbf{\delta}
\end{aligned}
\label{eq:trpo}
\end{equation}
where $L_{\mathbf{\theta}}(\mathbf{\theta}_{k})$ is the surrogate advantage, representing how good a policy $ \mathbf{\pi}_{\mathbf{\theta}}$ is with respect to an old policy $ \mathbf{\pi}_{\mathbf{\theta}_{k}}$:
\begin{equation}
L_{\mathbf{\theta}}(\mathbf{\theta}_{k}) = \underset{s,a\sim \mathbf{\pi}_{\mathbf{\theta}_{k}}}{E}\left[\frac{ \mathbf{\pi}_{\mathbf{\theta}}(a|s)}{ \mathbf{\pi}_{\mathbf{\theta}_{k}}(a|s)}A^{ \mathbf{\pi}_{\mathbf{\theta}_{k}}}(s,a)\right]
\label{eq:surrogate}
\end{equation}
where $A^{ \mathbf{\pi}_{\mathbf{\theta}_{k}}}$ is the advantage function. In the baselines implementation this advantage function is calculated using a value estimation (Actor-Critic structure)\cite{baselines}. \\
The analytical solution of the KL- divergence for each step is expensive, but is possible to approximate its value using a Taylor expansion of degree 2. To solve the optimization problem from Eq.\ref{eq:trpo}, its commonly used the Langrangian method. The obtained expression can be Taylor expanded, and the obtained result is known as the natural gradient \cite{kakade2001npg}. In order to solve analytically the natural gradient, the Fisher information matrix (FIM) needs to be calculated (the Hessian of the KL-divergence), which is not trivial to compute and store. For that TRPO uses a trick, optimizing a sub-problem, and finally performing a backtracking line search.\\
This approach, more than a structure, is way of optimizing the search of the parameters. Therefore it can be used with an Actor-Critc structure that calculates the value function and uses it in the advantage calculation for example.
\begin{algorithm}
\label{TRPO}
\caption{Trust Region Policy Optimization (TRPO)}
\begin{algorithmic}[1]
\STATE Input: initial policy parameters $\mathbf{\theta}_{0}$, initial value function parameters $\phi_{0}$
\STATE Hyperparameters: KL-divergence limit $\mathbf{\delta}$, backtracking coefficient $\mathbf{\alpha}$, maximum number of backtracking steps 
\FOR{ $k=0,1,2,...$}
\STATE Collect set of trajectories $\mathcal{D}_{k} = \{ \tau_{i}\}$ by running policy $\mathbf{\pi}_{k} =  \mathbf{\pi}(\mathbf{\theta}_{k})$ in the environment.
\STATE Compute rewards-to-go $\hat{R}_{t}$.
\STATE Compute advantage estimates, $\hat{A}_{t}$ (using any method of advantage estimation) based on the current value function $V_{\phi_{k}}$.
\STATE Estimate policy gradient as: \[\hat{g}_{k} = \frac{1}{\mathcal{D}_{k}} \sum_{\tau \in \mathcal{D}_{k}} \sum_{t = 0}^{T} \nabla_{\mathbf{\theta}} log  \mathbf{\pi}_{\mathbf{\theta}} (a_{t}| s_{t})|_{\mathbf{\theta}_{k}} \hat{A}_{t}.\]
\STATE Use the conjugate gradient algorithm to compute: \[\hat{x}_{k} \approx \hat{H}_{k}^{-1}\hat{g}_{k},\] where $\hat{H}_{k}^{-1}$ is the Hessian of the sample average KL-divergence.
\STATE Update the policy by backtracking line search with: \[\mathbf{\theta}_{k+1} = \mathbf{\theta}_{k} +  \mathbf{\alpha}^{j}\sqrt{\frac{2 \mathbf{\delta}}{\hat{x}_{k}^{T}\hat{H}_{k}\hat{x}_{k}}\hat{x}_{k}},\]
where $j \in \{0,1,2,...K\}$ is the smallest value which improves the sample loss and satisfies the sample KL-divergence constraint.
\STATE Fit value function by regression on mean-squared error: \[\phi_{k+1} = arg\,\min\limits_{\phi} \frac{1}{\mathcal{D}_{k}T}\sum_{\tau \in \mathcal{D}_{k}} \sum_{t = 0}^{T}(V_{\phi}(s_{t}) - \hat{R}_{t})^{2},\] typically via some gradient descent algorithm.
\ENDFOR
\end{algorithmic}
\end{algorithm}

\subsubsection{Proximal Policy Optimization (PPO)}
\label{methods:ppo}
Proximal Policy Optimization (PPO) is an alternative to Trust Region Policy Optimization (TRPO) \cite{schulman2015trust}, that attains data efficiency and reliable performance of TRPO while using first order optimization. In the case of standard PG methods, the gradient update is performed per data sample. On the other hand, PPO enables multiple epochs of mini-batch updates. There are a few variants of PPO in the literature, which optimize the "surrogate" objective or use adaptive KL penalty coefficient \cite{schulman2017proximal}. The Clipped Surrogate Objective is given as:
\begin{equation}
\begin{aligned}
L&^{clip}(\mathbf{\theta}) =\\ &\hat{E_{t}}[min(r_t(\mathbf{\theta})\hat{A_{t}}, clip(r_t(\mathbf{\theta}),1-\epsilon,1+\epsilon)\hat{A_{t}})]
\end{aligned}
\end{equation}
\noindent where $r_t(\mathbf{\theta}) = \frac{ \mathbf{\pi}_ \mathbf{\theta}(a_t|s_t)}{ \mathbf{\pi}_{\mathbf{\theta}_{old}}(a_t|s_t)}$, is the probability ratio of the current policy $ \mathbf{\pi}_ \mathbf{\theta}$ and the previous policy $ \mathbf{\pi}_{\mathbf{\theta}_{old}}$, $\hat{A_{t}}$ is an estimator of the advantage function at timestep $t$ and $\epsilon$ is a hyperparameter, for example $\epsilon = 0.2$. The first term is the "surrogate" objective that is also used in TRPO (Eq.\ref{eq:surrogate}). The second term, $clip(r_t(\mathbf{\theta}),1-\epsilon,1+\epsilon)\hat{A_{t}}$, is clipping the probability ratio, $r_t$, to be between the interval $[1-\epsilon,1+\epsilon]$. The $L^{clip}(\mathbf{\theta})$ takes the minimum of the un-clipped and clipped value, which excludes the change in the probability ratio when the objective improves and includes it when the objective is worse. The clipping prevents PPO from having a large policy update. The Adaptive KL Penalty Coefficient is an alternative to the clipped "surrogate" objective or an addition to it where the goal is to use the penalty on KL divergence and update the penalty coefficient to achieve some target KL divergence ($d_{targ}$) at each policy update. As described in \cite{schulman2017proximal}, the KL Penalty Coefficient performed worse than the surrogate objective, therefore the presented pseudocode and experimental evaluation of PPO uses the clipped surrogate objective.

\begin{algorithm}
\label{PPO}
\caption{Proximal Policy Optimization (PPO)}
\begin{algorithmic}[1]
\STATE Initialize the time steps ($T$)
\STATE Initialize the clipping value $\epsilon$ 
\FOR {$i = 1, N_{iterations}$}
\FOR {$t = 1, T$}
\STATE Run MLP policy and generate action $a_t$ 
\STATE Execute action $a_t$ in emulator and observe reward 
\STATE Update observation ($ob$) based on current joint positions and end-effector position 
\STATE Estimate advantage function $\hat{A_t} $ 
\FOR {$epoch=1,N_{epochs}$}
\STATE Compute SGD of loss function:
\begin{equation*}
\begin{aligned}
&L^{clip}(\mathbf{\theta}) = \\
&\hat{E_{t}}[min(r_t(\mathbf{\theta})\hat{A_{t}},clip(r_t(\mathbf{\theta}),1-\epsilon,1+\epsilon) \hat{A_{t}})]
\end{aligned}
\end{equation*}
\ENDFOR
\ENDFOR
\ENDFOR
\end{algorithmic}
\end{algorithm}

\subsubsection{Actor Critic using Kronecker-Factored Trust Region (ACKTR)}
\label{methods:acktr}
The idea of Actor Critic using Kronecker-Factored Trust Region (ACKTR) is to replace the Stochastic Gradient Descent (SGD), which explores the weight space inefficiently, and to optimize both the actor and the critic using Kronecker-factored approximate curvature (K-FAC) with trust region \cite{wu2017scalable}. ACKTR replaces SGD of A2C, the synchronous version of A3C \cite{mnih2016asynchronous}, and instead computes the natural gradient update. The natural gradient update is applied both to the actor and the critic.
 
ACKTR uses the K-FAC to compute the natural gradient update efficiently. In order to define a Fisher metric for RL policies, ACKTR uses a policy function that defines a distribution over actions given the current state, and takes the expectation over the trajectory distribution. The mathematical formulation for the Fisher metric is given by:

\begin{equation}
\label{eq:kfac_metric}
    F_a = \mathbb{E}_{p(\tau)}[\nabla_\mathbf{\theta}\log \mathbf{\pi}(a_t|s_t)(\nabla_\mathbf{\theta}\log \mathbf{\pi}(a_t|s_t))^T]
\end{equation}

\noindent where $p(\tau) = p(s_0) \prod^T_{t=0} \mathbf{\pi}(a_t|s_t)p(s_{t+1}|s_t,a_t)$ is the distribution of trajectories. In practice, we approximate the intractable expectation above with trajectories collected during training. In the case of training the critic, one can think of it as a least-squares function approximation problem. In this case, the most common second-order algorithm is Gauss-Newton, which approximates the curvature as the Gauss-Newton matrix $G:= \mathbb{E}[J^T J]$, where $J$ is the Jacobian mapping from parameters to outputs \cite{wright1999numerical}. The Gauss-Newton matrix is equivalent to the Fisher matrix for a Gaussian observation model, which allows to apply K-FAC to the critic as well. In more detail, the output of the critic $v$ is defined to be a Gaussian distribution with $p(v|s_t) \sim \mathcal{N}(v;V(s_t), \sigma^2)$. Setting $\sigma$ to 1 is equivalent to the vanilla Gauss-Newton method.

In the case when the actor and the critic are disjoint, it is possible to apply K-FAC updates to each of them using the same metric as defined in Equation \ref{eq:kfac_metric}. To prevent instability during training, it is important to use an architecture where the two networks both share lower-layer representations but have distinct output layers \cite{wang2016sample, mnih2016asynchronous}. The joint distribution of the policy and the value distribution can be defined by assuming independence of the two output distributions, for instance $p(a,v|s)= \mathbf{\pi}(a|s)p(v|s)$, and constructing the Fisher metric with respect to $p(a,v|s)$. This is similar to the standard K-FAC, except that we need to sample the two networks' outputs independently. In this case, the K-FAC to approximate the Fisher matrix is:

\begin{equation}
    F_v = \mathbb{E}_{p(\tau)}[\nabla\log p(a, v|s) \nabla\log p(a, v|s)^T]
\end{equation}

\noindent The pseudocode presented gives an overview of the ACKTR implementation used in our evaluation.

\begin{algorithm}
\label{ACKTR}
\caption{Actor Critic using Kronecker-Factored Trust Region (ACKTR)}
\begin{algorithmic}[1]
\STATE Assume shared parameter vector for the actor $a$ and $v$  for the critic. 
\STATE Assume global shared counter $T=0$
\STATE Initialize step counter $t \leftarrow 1$
\REPEAT
\STATE Reset action $a \leftarrow 0$ and state $s \leftarrow 0$
\STATE $t_{start} = t$
\STATE Get state $s_t$
\REPEAT
\STATE Perform action $a_t$ according to policy $ \mathbf{\pi}(a_t|s_t; \mathbf{\theta}^\prime)$
\STATE Receive reward $r_t$ and new state $s_{t+1}$
\STATE $t \leftarrow t+1$
\STATE $T \leftarrow T+1$
\UNTIL{terminal $s_t$ or $t-t_{start} == t_{max}$}
\STATE $R = \left\{
	       \begin{array}{ll}
      		 0,  \mathrm{for\; terminal\; s_t} \\
      		 V(s_t, \mathbf{\theta}^\prime_v), \mathrm{\ for\; non-terminal\; s_t.}
	       \end{array}
	     \right.$

\FOR{ $i=t-1,t_{start}$}
\STATE$R \leftarrow r_i +  \mathbf{\gamma} R$
\STATE Calculate natural gradient for the actor:
\STATE$F_a = \mathbb{E}_{p(\tau)}[\nabla_ \mathbf{\theta}\log \mathbf{\pi}(a_t|s_t)(\nabla_ \mathbf{\theta}\log \mathbf{\pi}(a_t|s_t))^T]$
\STATE$p(\tau) = p(s_0) \prod^T_{t=0} \mathbf{\pi}(a_t|s_t)p(s_{t+1}|s_t,a_t)$,
\STATE with $p(\tau)$ as distribution of trajectories collected during training
\IF{$a$ and $v$ are joint}
\STATE  Output of the critic $v$ is defined to be a Gaussian distribution: $p(v|s_t) \sim \mathcal{N}(v; V(s_t)\sigma^2)$
\STATE Apply Fisher matrix for the critic
\ENDIF
\IF{$a$ and $v$ are disjoint}
\STATE Apply K-FAC to approximate Fisher matrix for the critic  
\STATE $F_v = \mathbb{E}_{p(\tau)}[\nabla\log p(a, v|s) \nabla\log p(a, v|s)^T]$
\ENDIF
\ENDFOR
\UNTIL{$T > T_{max}$}
\end{algorithmic}
\end{algorithm}

\section{Experiments}
\label{experiments}

As previously presented by Zamora et al \cite{zamora2016extending}, for the benchmark experiments we use an extension of the OpenAI gym which is tailored for robotics. We added four additional environments to evaluate the algorithms, which match the modular MARA 6DoF. The environments differ on how they reward the actions taken, and are described in detail in gym-gazebo2 \cite{1903.06278}. For the training, we used the Gazebo simulator and corresponding ROS 2 packages, to convert the actions generated from each algorithm into appropriate trajectories that the robot can execute.

\begin{figure}[!h]
\centering
\includegraphics[width=0.48\textwidth]{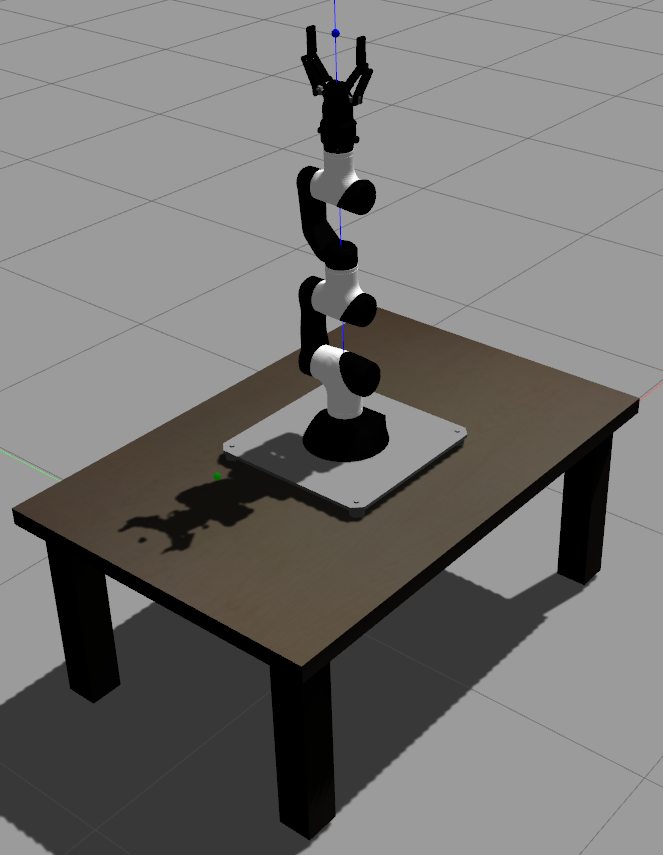}
\caption{gym-gazebo2 MARA robot environments displayed on Gazebo gzclient simulator. All environments are included since their differences are in how the learning is rewarded and not in the model}
\label{fig:enviroment}
\end{figure}

We set the initial position of the robot to zero for all joints and reset the robot to this initial position when the number of steps exceeds the maximum timesteps for an episode. We code this in an environment-specific variable denoted \textbf{\textit{max\_episode\_steps}}, which in our case is set to 2048. For these specific experiments, we located the fixed target at the coordinates $[x = -0.40028, y = 0.095615, z = 0.72466]$ with respect to the origin of the environment, which in our case is set to be the base of the 6DoF MARA robot; and the orientation at the quaternion $[w = 0., x = 0.7071068, y = 0.7071068, z = 0.]$, with respect to the table orientation. Each algorithm generates actions that are translated into the corresponding ROS 2 messages and are executed in simulation. The simulation then returns the observations (current joint positions and end-effector pose) and gives them to the algorithm. Figure \ref{fig:enviroment} illustrates the experimental environment. For each environment we perform one experiment consistent in a training for 1 million steps in the environment.

\begin{figure}[!h]
\centering
\includegraphics[width=0.48\textwidth]{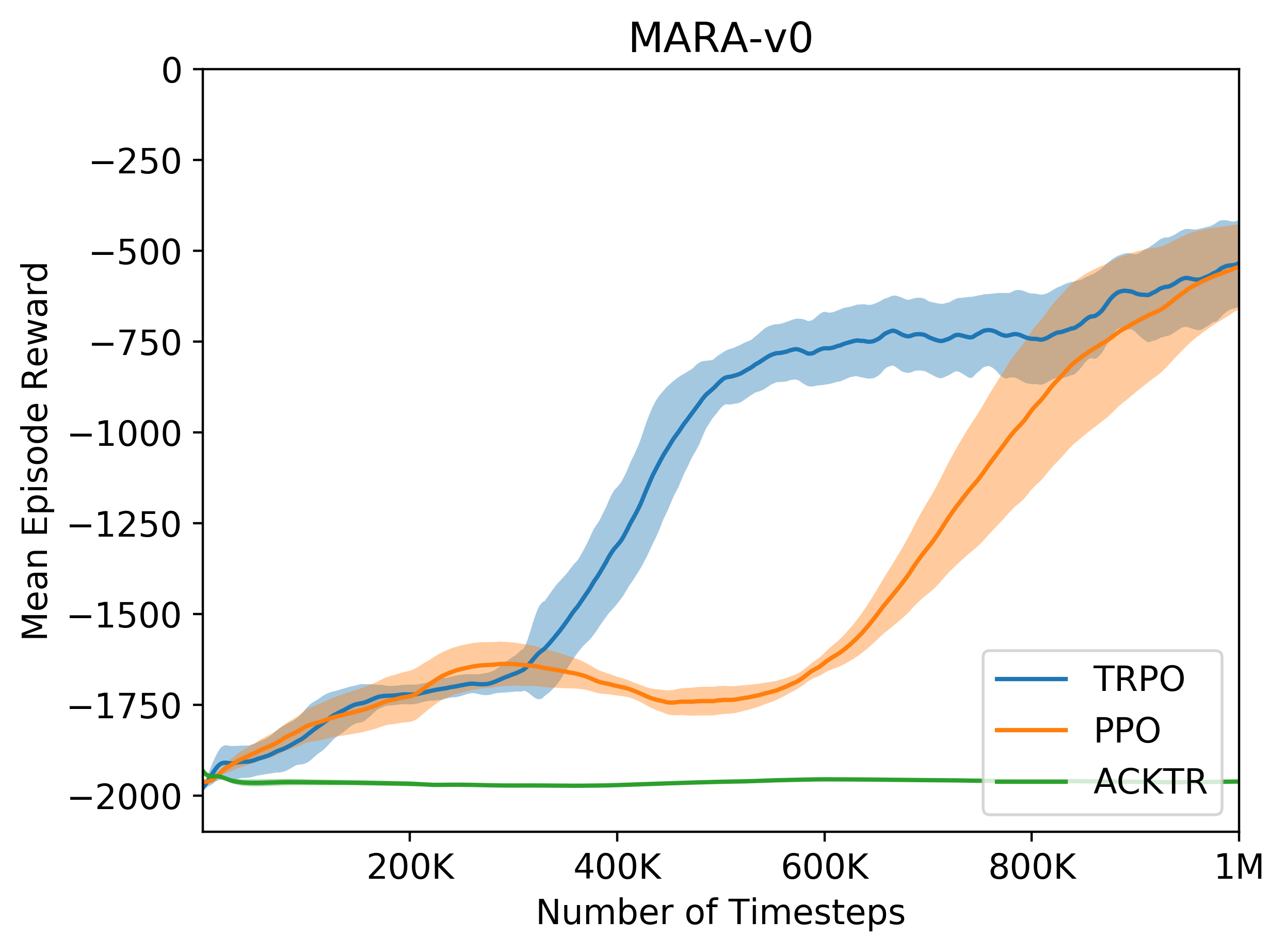}
\caption{Performance comparisons of the tested algorithms for the $MARA-v0$ environment. The shaded region denotes the deviation with respect to the previous 100 steps. PPO and TRPO seem to achieve a similar level of performance by the end of the experiments, even though TRPO seems to learn faster at earlier stages. The reward obtained by ACKTR remains unchanged compared with the other algorithms.}
\label{fig:MARA}
\end{figure}

\begin{figure}[!h]
\centering
\includegraphics[width=0.48\textwidth]{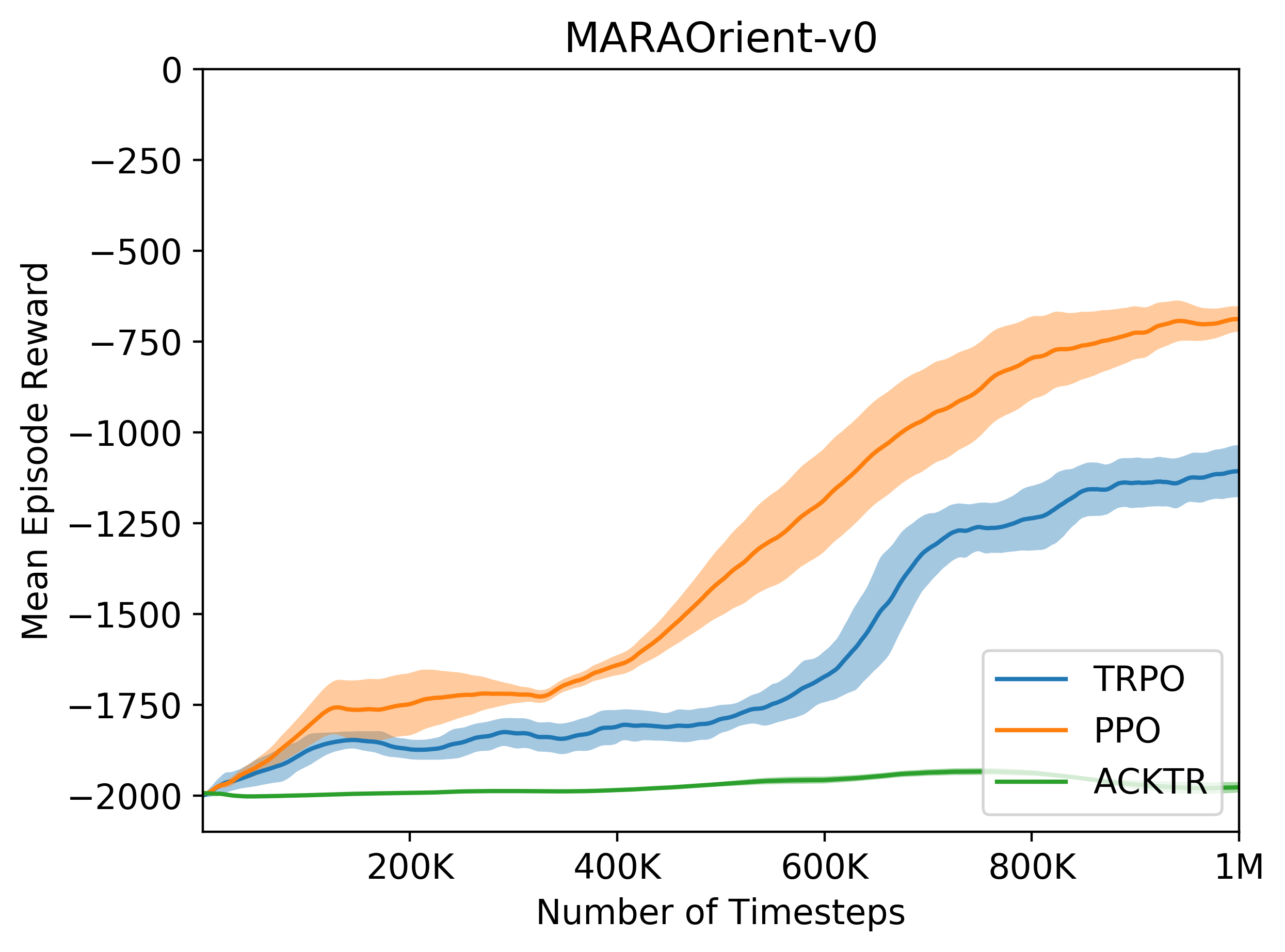}
\caption{Performance comparisons of the tested algorithms for the $MARAOrient-v0$ environment. The shaded region denotes the deviation of the rewards with respect to the previous 100 steps. PPO is able to get a slightly better result than TRPO in this environment, while ACKTR stays flat compared to the other two.}
\label{fig:MARAOrient}
\end{figure}

\begin{figure}[!h]
\centering
\includegraphics[width=0.48\textwidth]{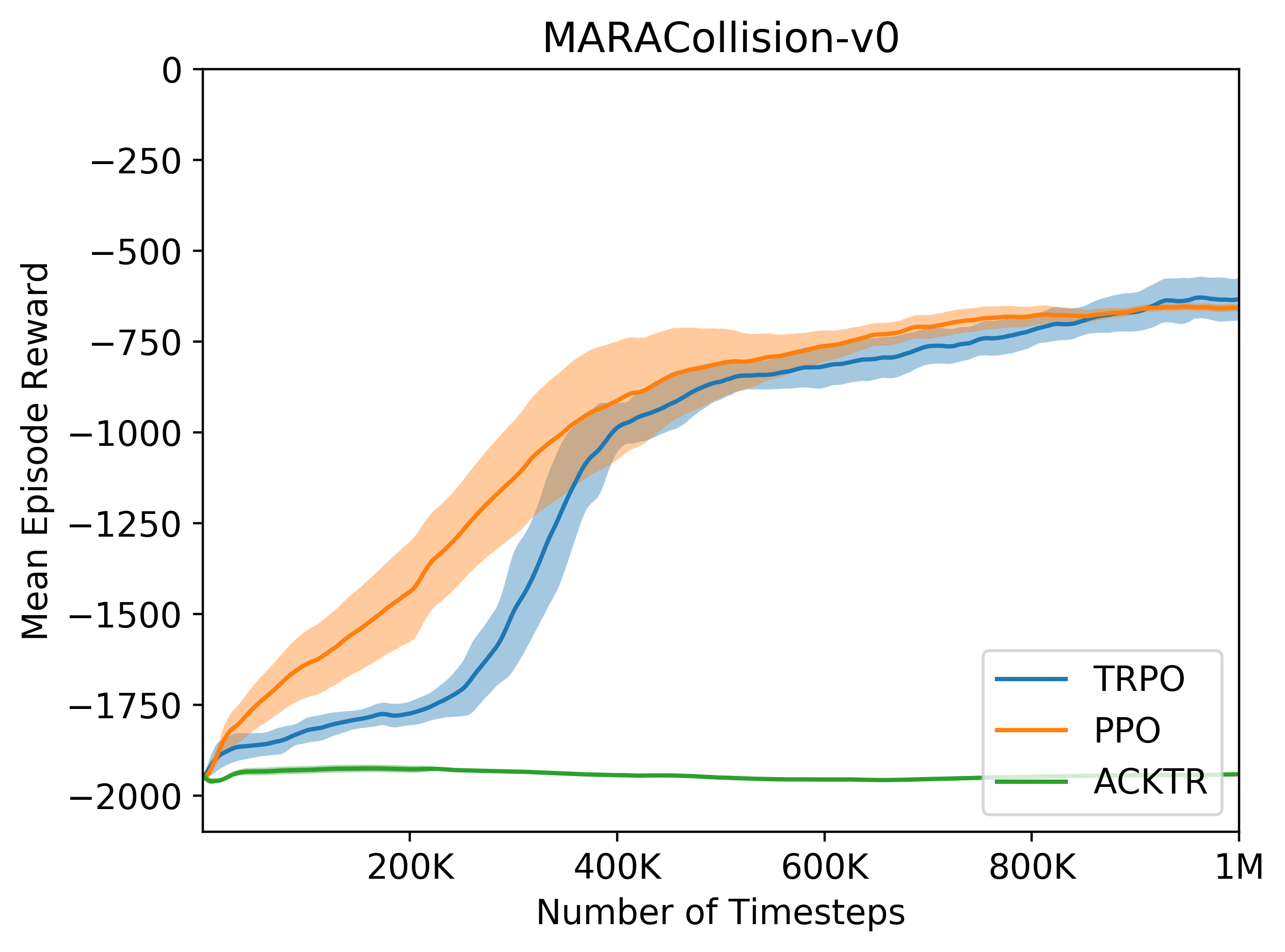}
\caption{Performance comparisons of the tested algorithms for the $MARACollision-v0$ environment trained. The shaded region denotes the deviation of the rewards with respect to the previous 100 steps. PPO and TRPO seem to have similar performance, while the reward obtained by ACKTR remains flat in comparison.}
\label{fig:MARACollision}
\end{figure}

\begin{figure}[!h]
\centering
\includegraphics[width=0.48\textwidth]{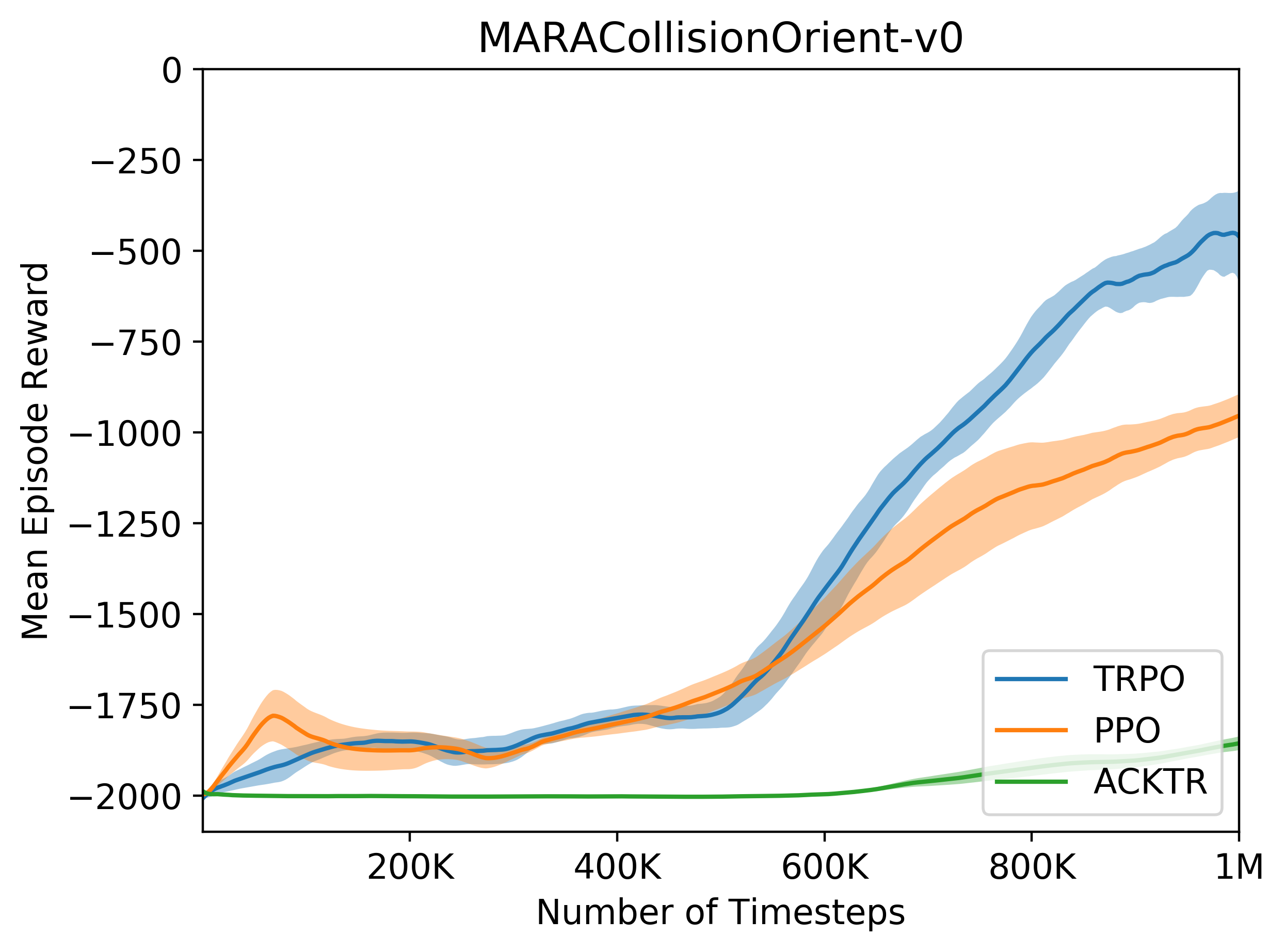}
\caption{Performance comparisons of the tested algorithms for the $MARACollisionOrient-v0$ environment. The shaded region denotes the deviation with respect to the previous 100 steps. TRPO shows better performance towards the end of the experiment compared to PPO, while this time, ACKTR shows some learning towards the end of the experiment, though not comparable with any of the other algorithms.}
\label{fig:MARACollisionOrient}
\end{figure}

Figure \ref{fig:MARA}, Figure \ref{fig:MARAOrient}, Figure \ref{fig:MARACollision} and Figure \ref{fig:MARACollisionOrient} show the reward obtained in the learning process for the different algorithms. In general TRPO and PPO show ability to learn at a similar pace, particularly in non-orient environments, which is not surprising given that they both have similar formulation. The discrepancies in orient environments might be due to the fact that those could be more dependant on the random initialization. ACKTR does not seem to be an efficient learner for this task. See more details in \cref{appendix}. It can be due to the used hyperparameters, which were all the same in the three different algorithms in order to compare them.

\section{Conclusion and Future work}
\label{conclusion}

We have presented evaluation of different DRL techniques for modular robotics. Our setup and framework consists of tools, such as ROS 2 and Gazebo, allowing a more realistic representation of the environment. Our results show that our proposed framework is stable during training of neural networks trough RL with policy-based methods.

There still remain many challenges within the DRL field for robotics. The main problems are the long training times, the simulation-to-real robot transfer, reward shaping, sample efficiency and extending the behaviour to diverse tasks and robot configurations. 

So far, our work with the modular robot MARA has focused on simple tasks such as reaching a point in space. In order to have an end-to-end training framework (from pixels to motor torques) and to perform more complex tasks, we aim to integrate additional rich sensory input such as vision. Inspired by the work of \cite{DuanASHSSAZ17, ho2016generative}, we intend to explore imitation learning that provides high-quality human training data through demonstrations which might be useful for the robot to learn to perform more complex tasks.

We envision the future of robotics to be modular robots where the trained network can generalize online to modifications in the robot such as change of a component or dynamic obstacle avoidance. In order to accomplish this, we aim to explore methods that allow novel training approaches of the robot for every new environment, type of robot or when the original task for which the network was trained for is changed. Inspired by \cite{abs-1710-09767}, we aim to evaluate meta-learning and hierarchical RL methods that allow to generalize to new tasks and environments by learning sub-policies; for instance motor primitives that can be reused across different sets of tasks, and even generalizing to unseen new tasks.

\appendix
\label{appendix}

\begin{figure}[!h]
\centering
\includegraphics[width=0.48\textwidth]{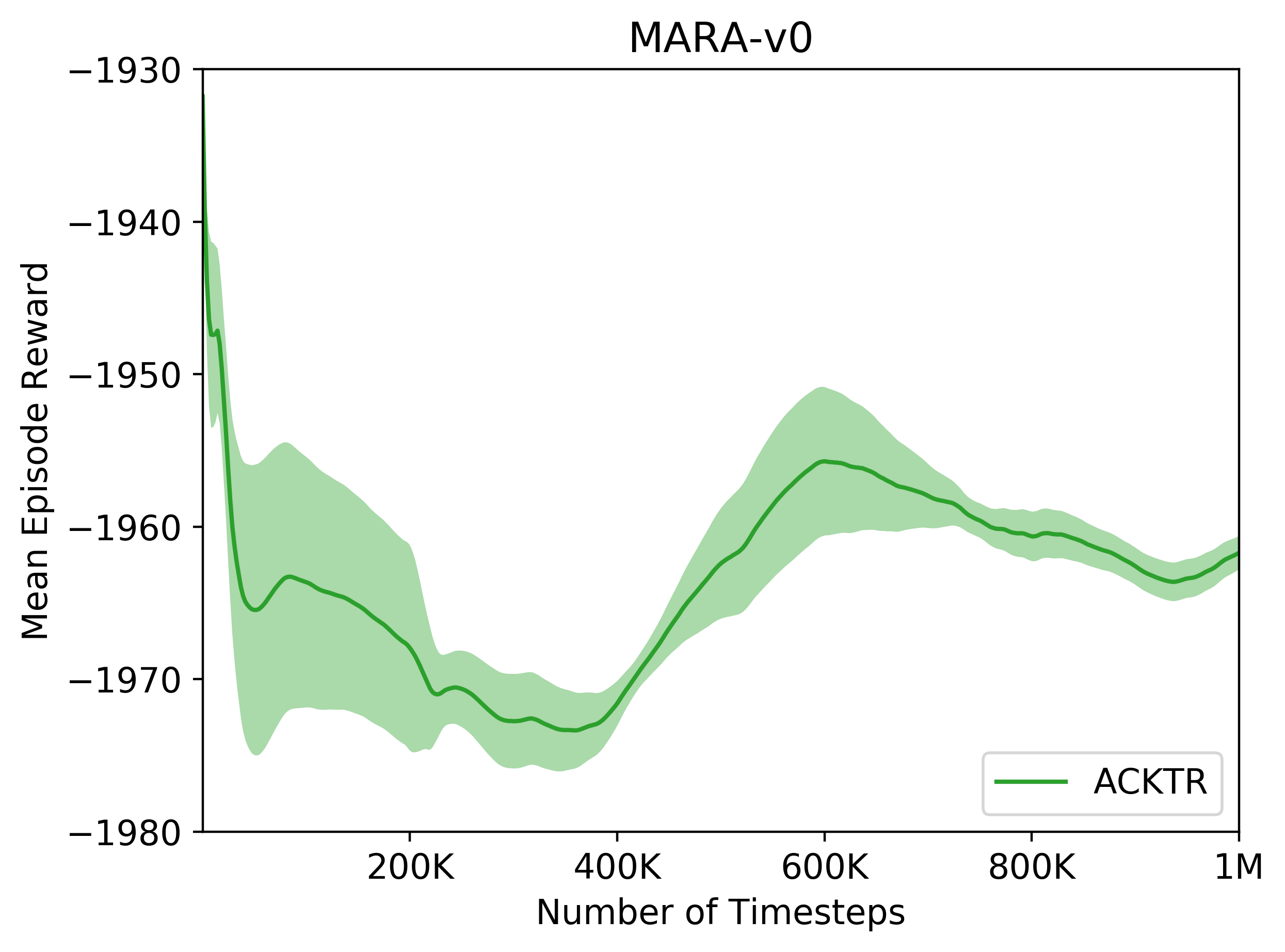}
\caption{Performance of $MARA-v0$ environment trained with ACKTR algorithm}
\label{fig:ACKTR-MARA}
\end{figure}

\begin{figure}[!h]
\centering
\includegraphics[width=0.48\textwidth]{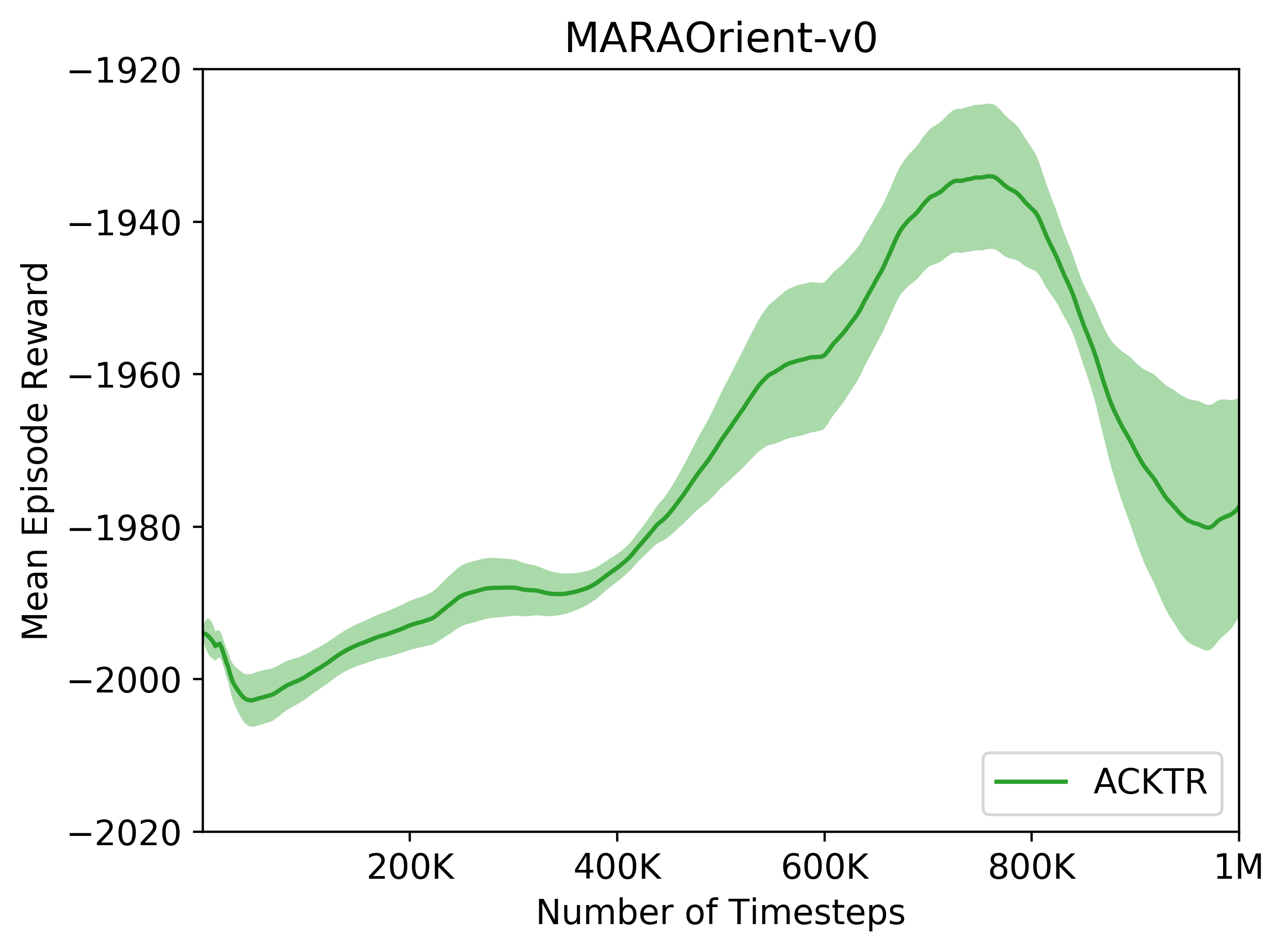}
\caption{Performance of $MARAOrient-v0$ environment trained with ACKTR algorithm}
\label{fig:ACKTR-MARAOrient}
\end{figure}

\begin{figure}[!h]
\centering
\includegraphics[width=0.48\textwidth]{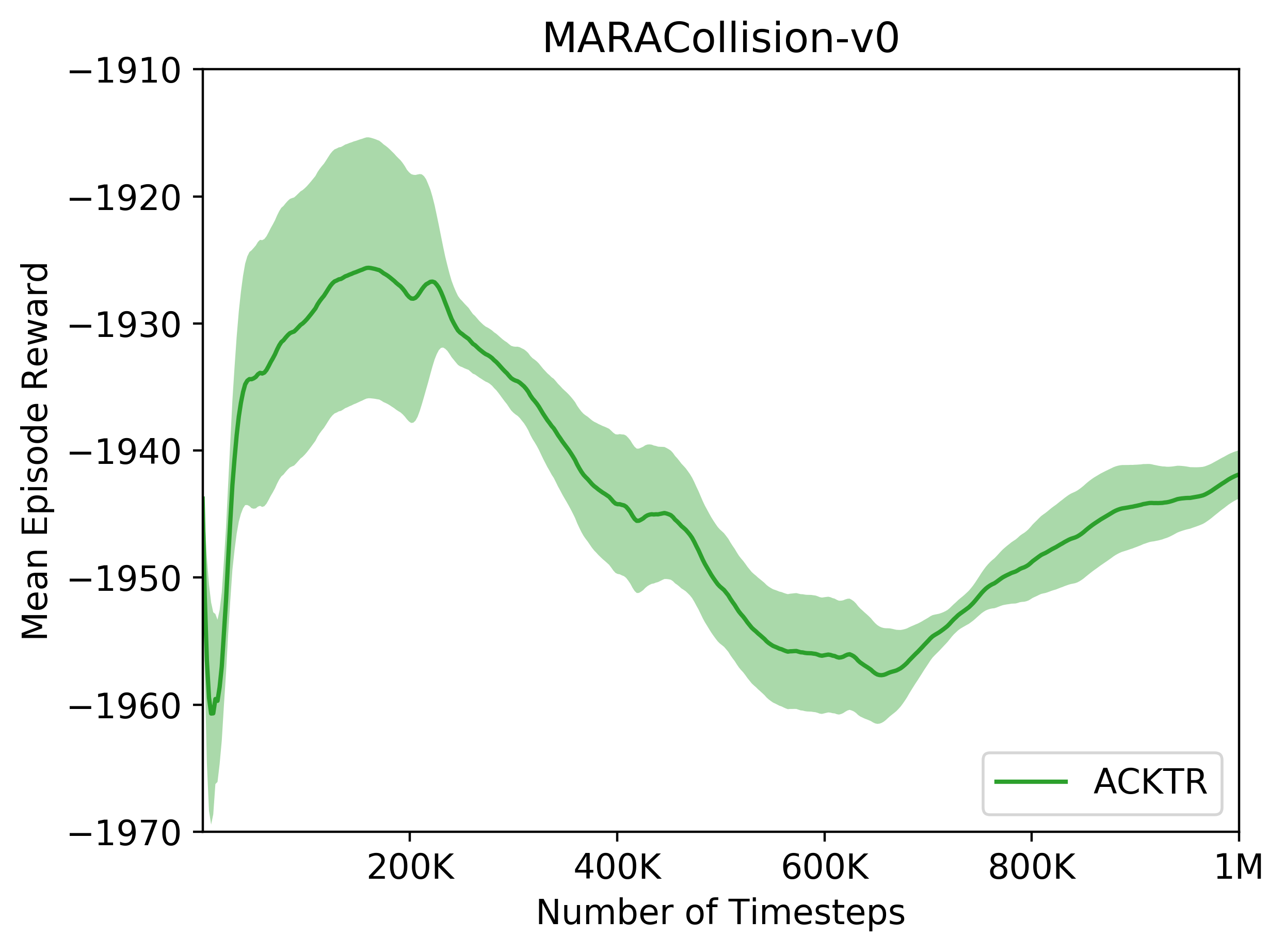}
\caption{Performance of $MARACollision-v0$ environment trained with ACKTR algorithm}
\label{fig:ACKTR-MARACollision}
\end{figure}

\begin{figure}[!h]
\centering
\includegraphics[width=0.48\textwidth]{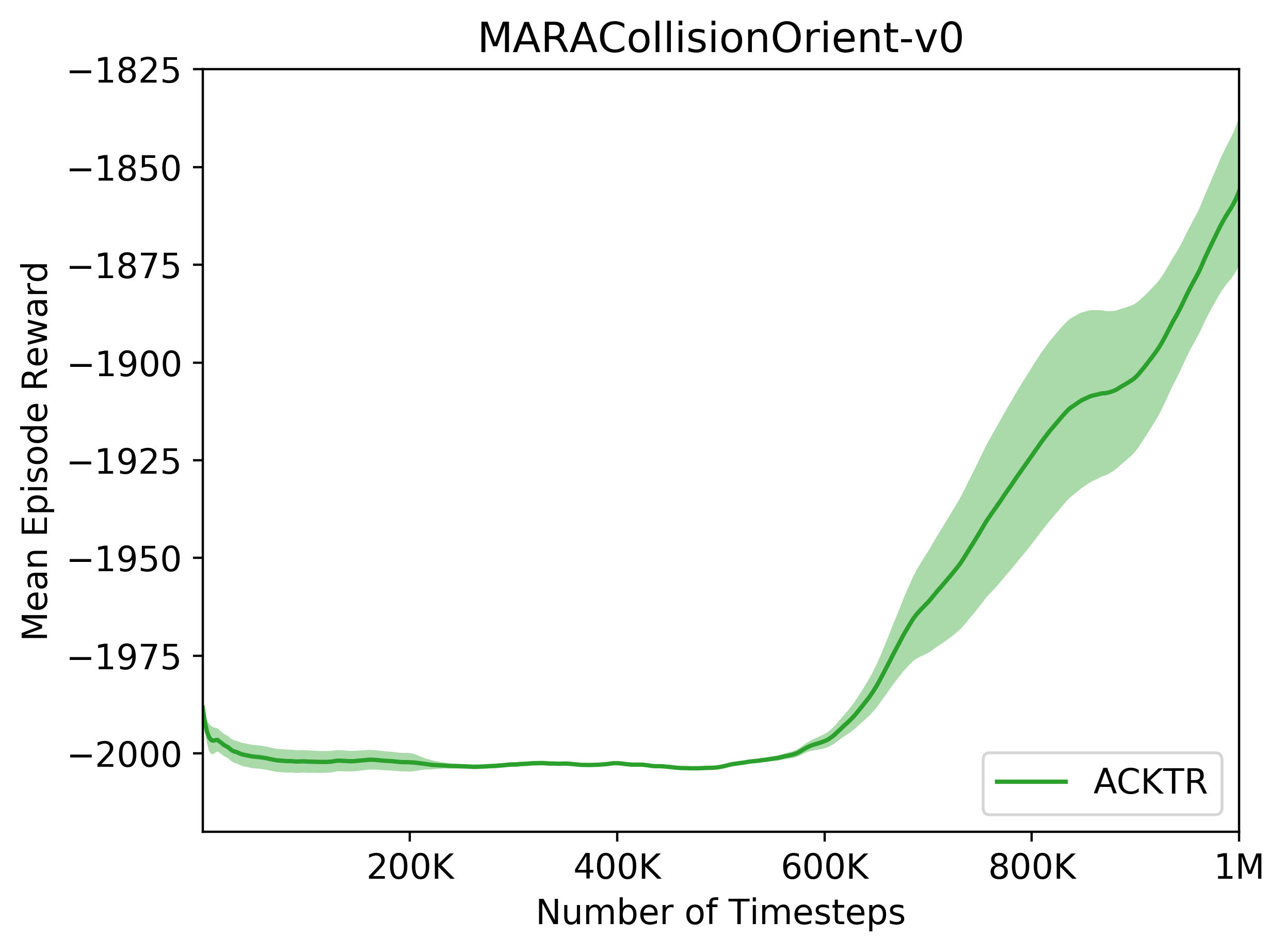}
\caption{Performance of $MARACollisionOrient-v0$ environment trained with ACKTR algorithm}
\label{fig:ACKTR-MARACollisionOrient}
\end{figure}

Figure \ref{fig:ACKTR-MARA}, Figure \ref{fig:ACKTR-MARAOrient}, Figure \ref{fig:ACKTR-MARACollision} and Figure \ref{fig:ACKTR-MARACollisionOrient} show the reward obtained in the learning process with ACKTR algorithm in each MARA robot environment. The shaded region denotes the deviation with respect to the previous 100 steps.

\ifCLASSOPTIONcaptionsoff
  \newpage
\fi





\bibliographystyle{IEEEtran}
\bibliography{references}

\begin{thebibliography}{10}
\providecommand{\url}[1]{#1}
\csname url@samestyle\endcsname
\providecommand{\newblock}{\relax}
\providecommand{\bibinfo}[2]{#2}
\providecommand{\BIBentrySTDinterwordspacing}{\spaceskip=0pt\relax}
\providecommand{\BIBentryALTinterwordstretchfactor}{4}
\providecommand{\BIBentryALTinterwordspacing}{\spaceskip=\fontdimen2\font plus
\BIBentryALTinterwordstretchfactor\fontdimen3\font minus
  \fontdimen4\font\relax}
\providecommand{\BIBforeignlanguage}[2]{{%
\expandafter\ifx\csname l@#1\endcsname\relax
\typeout{** WARNING: IEEEtran.bst: No hyphenation pattern has been}%
\typeout{** loaded for the language `#1'. Using the pattern for}%
\typeout{** the default language instead.}%
\else
\language=\csname l@#1\endcsname
\fi
#2}}
\providecommand{\BIBdecl}{\relax}
\BIBdecl

\bibitem{mayoral2018modular}
V.~Mayoral, R.~Kojcev, A.~Hern{\'a}ndez, I.~Zamalloa, and A.~Bilbao, ``Modular
  and self-adaptable (masa) strategy for building robots,'' in \emph{2018
  NASA/ESA Conference on Adaptive Hardware and Systems (AHS)}.\hskip 1em plus
  0.5em minus 0.4em\relax IEEE, 2018, pp. 90--95.

\bibitem{zamalloa2017dissecting}
I.~Zamalloa, R.~Kojcev, A.~Hern{\'a}ndez, I.~Muguruza, L.~Usategui, A.~Bilbao,
  and V.~Mayoral, ``Dissecting robotics-historical overview and future
  perspectives,'' \emph{arXiv preprint arXiv:1704.08617}, 2017.

\bibitem{DBLP:journals/corr/LevinePKQ16}
\BIBentryALTinterwordspacing
S.~Levine, P.~Pastor, A.~Krizhevsky, and D.~Quillen, ``Learning hand-eye
  coordination for robotic grasping with deep learning and large-scale data
  collection,'' \emph{CoRR}, vol. abs/1603.02199, 2016. [Online]. Available:
  \url{http://arxiv.org/abs/1603.02199}
\BIBentrySTDinterwordspacing

\bibitem{levine2013guided}
S.~Levine and V.~Koltun, ``Guided policy search,'' in \emph{Proceedings of the
  30th International Conference on Machine Learning (ICML-13)}, 2013, pp. 1--9.

\bibitem{peters2008reinforcement}
J.~Peters and S.~Schaal, ``Reinforcement learning of motor skills with policy
  gradients,'' \emph{Neural networks}, vol.~21, no.~4, pp. 682--697, 2008.

\bibitem{schulman2015trust}
J.~Schulman, S.~Levine, P.~Abbeel, M.~Jordan, and P.~Moritz, ``Trust region
  policy optimization,'' in \emph{Proceedings of the 32nd International
  Conference on Machine Learning (ICML-15)}, 2015, pp. 1889--1897.

\bibitem{schulman2017proximal}
J.~Schulman, F.~Wolski, P.~Dhariwal, A.~Radford, and O.~Klimov, ``Proximal
  policy optimization algorithms,'' \emph{arXiv preprint arXiv:1707.06347},
  2017.

\bibitem{wu2017scalable}
Y.~Wu, E.~Mansimov, R.~B. Grosse, S.~Liao, and J.~Ba, ``Scalable trust-region
  method for deep reinforcement learning using kronecker-factored
  approximation,'' in \emph{Advances in Neural Information Processing Systems},
  2017, pp. 5285--5294.

\bibitem{mnih2013playing}
V.~Mnih, K.~Kavukcuoglu, D.~Silver, A.~Graves, I.~Antonoglou, D.~Wierstra, and
  M.~Riedmiller, ``Playing atari with deep reinforcement learning,''
  \emph{arXiv preprint arXiv:1312.5602}, 2013.

\bibitem{todorov2012mujoco}
E.~Todorov, T.~Erez, and Y.~Tassa, ``Mujoco: A physics engine for model-based
  control,'' in \emph{Intelligent Robots and Systems (IROS), 2012 IEEE/RSJ
  International Conference on}.\hskip 1em plus 0.5em minus 0.4em\relax IEEE,
  2012, pp. 5026--5033.

\bibitem{7943603}
T.~Nogueira, S.~Fratini, and K.~Schilling, ``Autonomously controlling flexible
  timelines: From domain-independent planning to robust execution,'' in
  \emph{2017 IEEE Aerospace Conference}, March 2017, pp. 1--15.

\bibitem{zamora2016extending}
I.~Zamora, N.~G. Lopez, V.~M. Vilches, and A.~H. Cordero, ``Extending the
  openai gym for robotics: a toolkit for reinforcement learning using ros and
  gazebo,'' \emph{arXiv preprint arXiv:1608.05742}, 2016.

\bibitem{quigley2009ros}
M.~Quigley, K.~Conley, B.~Gerkey, J.~Faust, T.~Foote, J.~Leibs, R.~Wheeler, and
  A.~Y. Ng, ``Ros: an open-source robot operating system,'' in \emph{ICRA
  workshop on open source software}, vol.~3, no. 3.2.\hskip 1em plus 0.5em
  minus 0.4em\relax Kobe, 2009, p.~5.

\bibitem{koenig2004design}
N.~Koenig and A.~Howard, ``Design and use paradigms for gazebo, an open-source
  multi-robot simulator,'' in \emph{Intelligent Robots and Systems, 2004.(IROS
  2004). Proceedings. 2004 IEEE/RSJ International Conference on}, vol.~3.\hskip
  1em plus 0.5em minus 0.4em\relax IEEE, 2004, pp. 2149--2154.

\bibitem{baselines}
P.~Dhariwal, C.~Hesse, M.~Plappert, A.~Radford, J.~Schulman, S.~Sidor, and
  Y.~Wu, ``Openai baselines,'' \url{https://github.com/openai/baselines}, 2017.

\bibitem{Sutton:1999}
\BIBentryALTinterwordspacing
R.~S. Sutton, D.~McAllester, S.~Singh, and Y.~Mansour, ``Policy gradient
  methods for reinforcement learning with function approximation,'' in
  \emph{Proceedings of the 12th International Conference on Neural Information
  Processing Systems}, ser. NIPS'99.\hskip 1em plus 0.5em minus 0.4em\relax
  Cambridge, MA, USA: MIT Press, 1999, pp. 1057--1063. [Online]. Available:
  \url{http://dl.acm.org/citation.cfm?id=3009657.3009806}
\BIBentrySTDinterwordspacing

\bibitem{kullback1951information}
S.~Kullback and R.~A. Leibler, ``On information and sufficiency,'' \emph{The
  annals of mathematical statistics}, vol.~22, no.~1, pp. 79--86, 1951.

\bibitem{duan2016benchmark}
Y.~Duan, X.~Chen, R.~Houthooft, J.~Schulman, and P.~Abbeel, ``Benchmarking deep
  reinforcement learning for continuous control,'' in \emph{International
  Conference on Machine Learning}, 2016, pp. 1329--1338.

\bibitem{ghadir2017trpo}
\BIBentryALTinterwordspacing
A.~Ghadirzadeh, A.~Maki, D.~Kragic, and M.~Bj{\"{o}}rkman, ``Deep predictive
  policy training using reinforcement learning,'' \emph{CoRR}, vol.
  abs/1703.00727, 2017. [Online]. Available:
  \url{http://arxiv.org/abs/1703.00727}
\BIBentrySTDinterwordspacing

\bibitem{roboschool}
P.~Dhariwal, C.~Hesse, M.~Plappert, A.~Radford, J.~Schulman, S.~Sidor, and
  Y.~Wu, ``Openai baselines,'' \url{https://blog.openai.com/roboschool/}, 2017.

\bibitem{barrett2010transfer}
S.~Barrett, M.~E. Taylor, and P.~Stone, ``Transfer learning for reinforcement
  learning on a physical robot,'' in \emph{Ninth International Conference on
  Autonomous Agents and Multiagent Systems-Adaptive Learning Agents Workshop
  (AAMAS-ALA)}, 2010.

\bibitem{DBLP:journals/corr/RusuVRHPH16}
\BIBentryALTinterwordspacing
A.~A. Rusu, M.~Vecerik, T.~Roth{\"{o}}rl, N.~Heess, R.~Pascanu, and R.~Hadsell,
  ``Sim-to-real robot learning from pixels with progressive nets,''
  \emph{CoRR}, vol. abs/1610.04286, 2016. [Online]. Available:
  \url{http://arxiv.org/abs/1610.04286}
\BIBentrySTDinterwordspacing

\bibitem{DBLP:journals/corr/JamesJ16}
\BIBentryALTinterwordspacing
S.~James and E.~Johns, ``3d simulation for robot arm control with deep
  q-learning,'' \emph{CoRR}, vol. abs/1609.03759, 2016. [Online]. Available:
  \url{http://arxiv.org/abs/1609.03759}
\BIBentrySTDinterwordspacing

\bibitem{zhu2017target}
Y.~Zhu, R.~Mottaghi, E.~Kolve, J.~J. Lim, A.~Gupta, L.~Fei-Fei, and A.~Farhadi,
  ``Target-driven visual navigation in indoor scenes using deep reinforcement
  learning,'' in \emph{Robotics and Automation (ICRA), 2017 IEEE International
  Conference on}.\hskip 1em plus 0.5em minus 0.4em\relax IEEE, 2017, pp.
  3357--3364.

\bibitem{tobin2017domain}
J.~Tobin, R.~Fong, A.~Ray, J.~Schneider, W.~Zaremba, and P.~Abbeel, ``Domain
  randomization for transferring deep neural networks from simulation to the
  real world,'' \emph{arXiv preprint arXiv:1703.06907}, 2017.

\bibitem{Peters:2010}
J.~Peters, ``{P}olicy gradient methods,'' \emph{Scholarpedia}, vol.~5, no.~11,
  p. 3698, 2010, revision \#137199.

\bibitem{kakade2001npg}
\BIBentryALTinterwordspacing
S.~Kakade, ``A natural policy gradient,'' in \emph{Proceedings of the 14th
  International Conference on Neural Information Processing Systems: Natural
  and Synthetic}, ser. NIPS'01.\hskip 1em plus 0.5em minus 0.4em\relax
  Cambridge, MA, USA: MIT Press, 2001, pp. 1531--1538. [Online]. Available:
  \url{http://dl.acm.org/citation.cfm?id=2980539.2980738}
\BIBentrySTDinterwordspacing

\bibitem{mnih2016asynchronous}
V.~Mnih, A.~P. Badia, M.~Mirza, A.~Graves, T.~Lillicrap, T.~Harley, D.~Silver,
  and K.~Kavukcuoglu, ``Asynchronous methods for deep reinforcement learning,''
  in \emph{International Conference on Machine Learning}, 2016, pp. 1928--1937.

\bibitem{wright1999numerical}
S.~J. Wright and J.~Nocedal, ``Numerical optimization,'' \emph{Springer
  Science}, vol.~35, no. 67-68, p.~7, 1999.

\bibitem{wang2016sample}
Z.~Wang, V.~Bapst, N.~Heess, V.~Mnih, R.~Munos, K.~Kavukcuoglu, and
  N.~de~Freitas, ``Sample efficient actor-critic with experience replay,''
  \emph{arXiv preprint arXiv:1611.01224}, 2016.

\bibitem{1903.06278}
N.~G. Lopez, Y.~L.~E. Nuin, E.~B. Moral, L.~U.~S. Juan, A.~S. Rueda, V.~M.
  Vilches, and R.~Kojcev, ``gym-gazebo2, a toolkit for reinforcement learning
  using ros 2 and gazebo,'' 2019.

\bibitem{DuanASHSSAZ17}
\BIBentryALTinterwordspacing
Y.~Duan, M.~Andrychowicz, B.~C. Stadie, J.~Ho, J.~Schneider, I.~Sutskever,
  P.~Abbeel, and W.~Zaremba, ``One-shot imitation learning,'' in \emph{Advances
  in Neural Information Processing Systems 30: Annual Conference on Neural
  Information Processing Systems 2017, 4-9 December 2017, Long Beach, CA,
  {USA}}, 2017, pp. 1087--1098. [Online]. Available:
  \url{http://papers.nips.cc/paper/6709-one-shot-imitation-learning}
\BIBentrySTDinterwordspacing

\bibitem{ho2016generative}
J.~Ho and S.~Ermon, ``Generative adversarial imitation learning,'' in
  \emph{Advances in Neural Information Processing Systems}, 2016, pp.
  4565--4573.

\bibitem{abs-1710-09767}
\BIBentryALTinterwordspacing
K.~Frans, J.~Ho, X.~Chen, P.~Abbeel, and J.~Schulman, ``Meta learning shared
  hierarchies,'' \emph{CoRR}, vol. abs/1710.09767, 2017. [Online]. Available:
  \url{http://arxiv.org/abs/1710.09767}
\BIBentrySTDinterwordspacing

\end{thebibliography}

\end{document}